\documentclass[sigconf]{aamas}
\usepackage{balance} % for balancing columns on the final page
\usepackage{latexsym}
\usepackage{physics}
\usepackage{amsmath}
\usepackage{tikz}
\usepackage{mathdots}
\usepackage{cancel}
\usepackage{color}
\usepackage{siunitx}
\usepackage{array}
\usepackage{multirow}
\usepackage{gensymb}
\usepackage{tabularx}
\usepackage{booktabs}
\usetikzlibrary{fadings}
\usetikzlibrary{patterns}
\usetikzlibrary{shadows.blur}
\usetikzlibrary{shapes}
\usepackage{svg}

\usepackage{lipsum}

\setcopyright{ifaamas}
\acmConference[AAMAS '21]{Proc.\@ of the 20th International Conference on Autonomous Agents and Multiagent Systems (AAMAS 2021)}{May 3--7, 2021}{Online}{U.~Endriss, A.~Now\'{e}, F.~Dignum, A.~Lomuscio (eds.)}
\copyrightyear{2021}
\acmYear{2021}
\acmDOI{}
\acmPrice{}
\acmISBN{}

\acmSubmissionID{28}

\title[AAMAS-2021 Cognitive Homeostatic Agents]{Cognitive Homeostatic Agents}
\subtitle{Blue Sky Ideas Track}

\author{Amol Kelkar}
\affiliation{
  \institution{i3ai.org}}
\email{amol@i3ai.org}

\begin{abstract}
Human brain has been used as an inspiration for building autonomous agents, but it is not obvious what level of computational description of the brain one should use. This has led to overly opinionated symbolic approaches and overly unstructured connectionist approaches. We propose that using homeostasis as the computational description provides a good compromise. Similar to how physiological homeostasis is the regulation of certain homeostatic variables, cognition can be interpreted as the regulation of certain 'cognitive homeostatic variables'. We present an outline of a Cognitive Homeostatic Agent, built as a hierarchy of physiological and cognitive homeostatic subsystems and describe structures and processes to guide future exploration. We expect this to be a fruitful line of investigation towards building sophisticated artificial agents that can act flexibly in complex environments, and produce behaviors indicating planning, thinking and feelings.
\end{abstract}

\keywords{artificial agent;
homeostasis;
cognitive homeostasis;
hierarchical reinforcement learning;
predictive coding;
free energy principle;
cognitive homeostatic agent; robotic control; lifelong learning; online learning; emotions; rewards; predictive processing; hippocampus; engrams}

\newcommand{\BibTeX}{\rm B\kern-.05em{\sc i\kern-.025em b}\kern-.08em\TeX}

\begin{document}

%%% The following commands remove the headers in your paper. For final 
%%% papers, these will be inserted during the pagination process.

\pagestyle{fancy}
\fancyhead{}

%%% The next command prints the information defined in the preamble.

\maketitle 

%%%%%%%%%%%%%%%%%%%%%%%%%%%%%%%%%%%%%%%%%%%%%%%%%%%%%%%%%%%%%%%%%%%%%%%%
\section{Introduction}

The principle goal of artificial agents is intelligent behavior\cite{russell2009artificial}. Early research towards building Artificial Intelligent (AI) systems viewed human cognition as analogous to symbolic computation in digital computers\cite{Garson1991}. Later, connectionist approaches used artificial neural networks and attempted to implement cognitive features by mapping them to functions, taking advantage of the universal approximation theorem\cite{csaji2001approximation}. Recently, deep neural networks, reinforcement learning, meta-learning and related research areas have achieved great success in many areas of AI, such as playing console games\cite{mnih2013atari-rl}, robotic object manipulation\cite{openai2019solving}, natural language understanding and generation\cite{brown2020language}, and so on. Despite these advances, many challenges remain\cite{dulac2019challenges} for building robust artificial agents that can act in real world environments, interacting with humans and other agents. First, such agents must adapt continuously to changes in their environment, which requires online and lifelong learning. That remains difficult\cite{chen2018lifelong} for current deep learning systems for reasons such as catastrophic forgetting\cite{mccloskey1989catastrophic}. Second, current artificial agents are driven by arbitrarily defined external rewards, or rewards derived from unsatisfactory signals such as novelty. Third, affects are important signals\cite{scheutz2012affect} that may help the agents deal with complex interactions with their environment, other agents and with humans. This area remains under-explored. Lastly, mortality and the life regulation to postpone death are some of the fundamental driving forces behind human creativity, industriousness and morality. It is unclear whether artificial agents would achieve those attributes without the meta objective of self-preservation \cite{Man2019}. 

The computational description of the brain that symbolic and connectionist approaches use are over and under constrained, respectively. To investigate what an appropriate level of computational description would be, we go back to the first principles and explore how intelligence evolved and operates in living organisms.

% In nature, intelligent behavior is a byproduct of life regulation processes, which aren't aimed at producing intelligence\cite{downing2015intelligence}. Can we build systems that are not explicitly designed to produce intelligent behavior and allow intelligence to emerge? What computations would such a system implement? What motivations would such a system operate on? Would such systems produce higher cognitive behaviors?

\section{Homeostasis and Cognition}

Adaptive behavior is a hallmark of all living organisms. Single-celled organisms such as bacteria can move towards food and away from noxious chemicals\cite{alon1999robustness}, and routinely make complex decisions as part of bacterial colonies\cite{ross2014collective}. Plants share resources\cite{simard1997net} with each other through networks of fungi. Social insects lead complex, socially co-dependent lives\cite{oster1978caste}. Birds can count\cite{emery2006cognitive} and dogs can understand many verbal commands\cite{coren1994intelligence}. Octopuses can solve puzzles and remember large underwater terrains using landmarks\cite{godfrey2016other}. It is evident that there is a continuum of cognitive capabilities across living organisms. Humans have exceptionally sophisticated versions of some of these capabilities, but the underlying physiological processes seem to be mostly unremarkable compared to other evolutionarily proximate animal species. Building intelligent agents based on our shared evolutionary inheritance is an under-explored approach, which we focus on in this article.

Adaptive behavior of living organisms is grounded on an internal reward system \cite{deci2010intrinsic, barto2013intrinsic} that is tuned to their homeostatic imperatives \cite{Morville2018}. Deviation (or the prediction of deviation) of a homeostatic variable has a positive or negative valence\cite{Joffily2013} based on whether the deviation is towards or away from a homeostatic target state, i.e. whether the deviation is beneficial or harmful for the organism. For example, on a cold winter night, core body temperature starts to fall, moving away from nominal body temperature, which feels bad. Wearing a sweater feels good because it helps move body temperature closer towards the target value. These valences lead to affects and feelings\cite{Craig2002}, and drive behavior. Can homeostasis then be a good candidate computational description for building artificial agents that can act in complex environments in a grounded manner?

\subsection{(Re)defining Homeostasis}
Although the term homeostasis is often used to represent a narrow set of reactionary behaviors that move an organism towards a fixed set point, Cannon\cite{Cannon1929} originally intended the term to represent myriad physiological strategies to sense, maintain and defend regulated variables, as "wisdom of the body". This broader usage includes allostasis\cite{sterling1988allostasis}, the learned anticipatory responses and non-stationary target states, and other related mechanisms such as predictive homeostasis \cite{Moore-Ede1986}
% , homeorheusis \cite{Nicolaidis2011}, homeorhesis \cite{Waddington1968} and so on
. We use the term in its broadest scope, as explored under the free energy principle\cite{Friston2013-life-as-we-know-it}, where homeostasis includes all processes that keeps an organism within a relatively small set of favourable states that are conducive to its ongoing existence\cite{Corcoran2017}. Under this definition, homeostasis not only regulates near term physiological states of the organism and survival, but also regulates well-being and long term flourishing.

% By this definition of homeostasis, for relatively simple organisms without nervous systems, all behavior is motivated by homeostasis. Such organisms do not have the ability to ignore homeostatic drives and this 'Embodied Agency' is sufficient to explain away behavior in such organisms. On the other hand, a remarkable property of organisms with nervous systems is their ability to apparently resist homeostatic drives and act towards longer term goals. We propose that what appears as resisting homeostatic drives is in reality following homeostasis of higher order variables. Further, we postulate that higher cognitive functions such as thinking and logical reasoning are byproducts of the process of homeostasis of yet higher order variables.

\subsection{Hierarchical Homeostatic Systems}

Each organism is built as a hierarchy of homeostatic systems\cite{Friston2013-life-as-we-know-it,Kirchhoff2018,Connolly2017} such as membranes, sub-cellular organelles, cells, organs and organ systems.  Such hierarchies of homeostatic systems remain stable only under specific conditions. At every level in the hierarchy, a controller subsystem must act to ensure their own homeostasis, the homeostasis of all subordinate subsystems and the homeostasis of the combined system. To achieve this, controllers need to have their own homeostatic variables ("hvars"), and hvars to represent the homeostatic state of subordinate subsystems, conveyed by bottom-up, interoceptive signals from those subsystems. Top-down signals allow a controller to influence homeostasis of subordinate subsystems. These top-down signals convey the expectations of future interoceptive signals\cite{paulus2006insular}, which imply changes to steady state, i.e. target values, of certain hvars for the subsystems. The changed hvar targets prompt the subsystems to initiate appropriate actions or generate expectations for their own subsystems. This activity cascade\cite{friston2019waves} results in the engagement of the entire hierarchy towards meeting the top-down expectations. 
For example, when bacteria or viruses invade the body and cause tissue injury, one of the immune system’s responses is to produce pyrogens\cite{pyrogen}. These chemicals are carried by the blood to the brain, where they affect the functioning of the hypothalamus by effectively increasing the target value for the core body temperature hvar. In response, the hypothalamus raises the body’s temperature above the normal range, thereby causing a fever. 
% Note that although hierarchy implies a strict tree-like organization, that is not a requirement and in general, the dependencies follow a cyclic directed graph.

Signalling among subsystems is critical for a hierarchical homeostatic system to operate. Many organisms and bodily subsystems use chemical messengers\cite{taylor2010top}. It is difficult to convey different types of messages over long distances quickly and at precise targets using chemical signalling. On the other hand, neurons carry signals relatively rapidly and precisely and networks of neurons are capable of performing computations. This allows a nervous system based homeostatic controller to receive homeostatic state updates from a large number of subsystems and send back control signals to many subsystems, resulting in a fine grained mechanism to maintain homeostasis. This provides a clue why, in  animals species, the nervous system took up the role as the highest level homeostatic controller.

\subsection{Brain as a Homeostatic Controller}
Over evolutionary time scales, nervous systems have proven to be highly advantageous. Neural networks can represent homeostatic state information and other useful sensory information, and complex computations can be performed across signals from different modalities to produce highly adaptive control signals. With a nervous system, organisms can optimize actions over longer time horizons, which sometimes requires ignoring or acting against immediate homeostatic drives.

This ability of a nervous system based controller to potentially drive subsystems against their homeostatic imperatives may prove dangerous for the organism. First, subsystems may be driven to states where they are no longer able to maintain their homeostasis, leading to organ failure. Second, the nervous system operates on partial information and may make decisions that are not optimal. Third, the nervous system's control algorithms, either genetically specified or learned, may be insufficient to handle the current situation. For a stable homeostatic system hierarchy, it is imperative that controllers must treat bottom-up signals conveying subsystem homeostatic state, such as a painful sensation, as valenced, i.e. conveying a "good" or "bad" feeling. The controllers must consider these signals as "privileged", compared to other signals that may carry miscellaneous information. Controllers must pay a high metabolic cost if they choose to ignore these privileged signals. For example, ignoring hunger while voluntary fasting requires constant upkeep of the resolve to fast. Note that although privileged signals are present at every level in the hierarchy, they may become "felt" only at certain systems in the brain.

\section{Cognitive Homeostasis}
Cognitive science typically makes a sharp distinction between mental activity like planning and activity for physiological homeostasis. We propose that what appears as resisting homeostatic drives during planning is in reality following homeostasis of other, higher order variables.

The brain is organized roughly as a hierarchy of subsystems\cite{Hilgetag2020}. One question relevant for this discussion is whether the brain represents a single homeostatic system or a hierarchy of homeostatic subsystems. Note that a single homeostatic system can be organized internally as a hierarchy of subsystems, but in that case, the subsystems do not represent homeostatic variables of their own and thus act merely as functional modules. Looking at what hvars the brain represents will help answer this question.

\subsection{Cognitive Homeostatic Variables}
As a homeostatic regulator, the brain must represent the physiological hvars such as the core body temperature, sodium levels, blood pressure, heart rate and so on. Basic emotions\cite{BasicEmotions2017} such as fear, anger, joy, sadness, surprise and disgust can be interpreted as a set of hvars because the brain attempts to regulate them. For example, fear is the expectation of future harm and can be interpreted as an hvar whose steady state is the expectation of ongoing safety. Interpreting emotions as hvars is appropriate because emotions carry valence and by definition, valence is available only when a quantity affects homeostasis of the organism. More complex attributes such as the desire for well-being is also a homeostatic variable, because it has a valence specific to an individual. Long term goals are hvars because achieving those goals has a positive valence. Habits are hvars. Even short term goals such as raising a hand can be cast as a temporary hvar. Successful completion of every thought has a positive valence, while jumbled, incoherent thoughts that do not lead to a resolution have a negative valence. As a result, each thought can also be interpreted as a process that leads to the target state of a temporary hvar. Indeed, all facets of cognition that have valence can be cast as homeostasis via corresponding hvars. We call these presumed hvars Cognitive Homeostatic Variables ("cognitive hvars"). 
% Note that some cognitive processes are not mediated via cognitive hvars, for instance sensory and motor control, because those do not have any valence. Such processes represent utility machinery that aids in the regulation of some cognitive hvars.

Cognitive hvars may be fleeting like those for thoughts, short term like those for intentional actions, long term like a new year's resolution, or lifelong like career goals or social standing. They may be genetically specified, such as parenting instincts, or learned, such as habits.

\subsection{Why Cognitive Homeostasis?}
Why re-frame entire cognition as homeostasis? First, doing so gives a normative view of the behavior of all organisms, from single celled organism to humans, and potentially all self-organized systems such as societies and bacterial colonies. Second, it allows for the prioritization across physiological drives and cognitive goals. For example, conflicting goals such as energy homeostasis (hunger) and the desire to fast (not eat) can influence behavior from different levels in the homeostatic hierarchy, and actions result based on the relative influence of the two hvars and their interplay with other hvars in the system. Third, evolution often results in the reuse of components and patterns. In this case, it is conceivable that the neural patterns useful for physiological homeostasis were repeated for cognitive homeostasis as well, which presents an approach to investigate the evolution of nervous systems. Forth, this unification opens up the possibility of new ways of explaining free-will, self-hood and consciousness.

\subsection{Modelling Cognitive HVars}
Can we find a common computational description of how hvars are implemented in biological systems, so it can be applied in the construction of artificial agents? In nature, the mechanisms used for homeostasis vary substantially. For example, in the mitochondrial membrane, proton gradient across the membrane is maintained and is thus a homeostatic variable. When proton gradient shifts away from desired levels, molecular machines called proton pumps are triggered to restore the gradient. Locusts possess genetically specified visual neurons (the descending contralateral movement detectors, DCMDs) that detect predatory birds' looming motion\cite{LocustSanter2012} and trigger escape jumps. In this case, the hvar is implemented as specific afferent connectivity pattern of the DCMDs. Core body temperature hvar of endothermic animals is implemented as a system of sensors, neural and chemical signalling and several populations of neurons.
 
All these mechanisms can be summarized computationally as a set of desirable trajectories described using partial state\cite{senft2017toward} of the system. A trajectory is a sequence of changes described using certain partial state of the system over $\tau$ time steps, like a short animation clip. An hvar comprises the set of such trajectories that use the same set of partial state features. For example, thermoregulation hvar is a set of trajectories where starting state indicates a system that is either too cold or too warm, and ending state is where the system is closer to the optimal temperature. Other state features of the system, say blood glucose level, are not relevant to describe these trajectories, thus "partial state" above. 

In other words, an hvar is a human-understandable interpretation of a set of trajectories. In a cognitive homeostatic system, we may not have interpretation for all trajectories. For this reason, when building such a system, we work directly with trajectories and not hvars.

\section{Cognitive Homeostatic Agents}
We propose building Cognitive Homeostatic Agents, artificial agents that are based on the Cognitive Homeostasis principles.
\begin{figure}[t!]
  \centering
  \resizebox{1.0\linewidth}{!}{
  \includegraphics{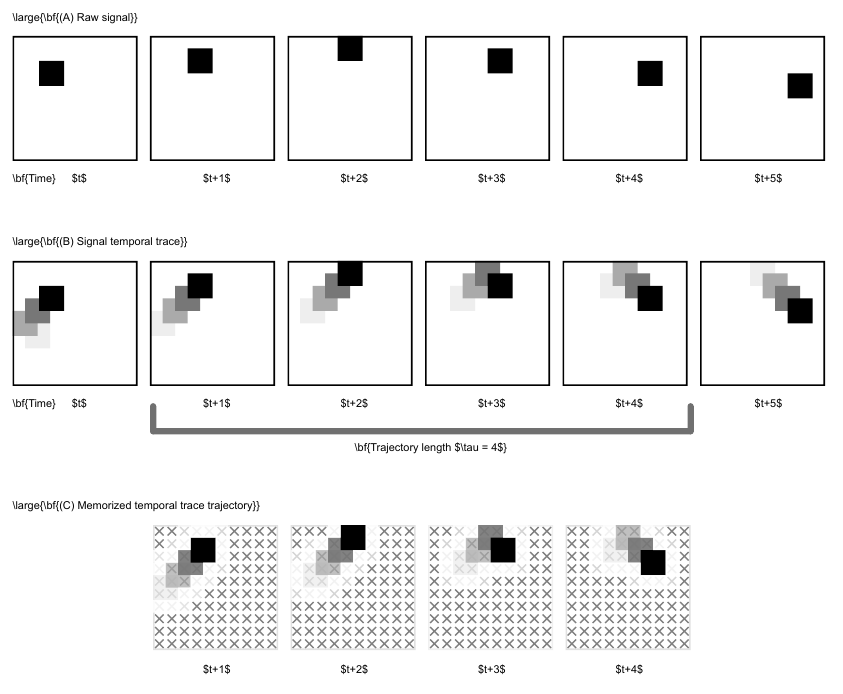}}
%   \includesvg{trajectory.svg}}
  \caption{Temporal Trace Trajectory. (A) shows a signal generated from a signal source such as a sensory modality. (B) shows temporal trace, i.e. motion blurred, version of the signal. Temporal Trace Trajectory is a $\tau$ time step sequence of the signal's temporal trace. When storing in memory, trajectories can be memorized using a partial state space, indicated by the opaque areas in (C).}
  \label{figure:ttt}
\end{figure}
% \subsection{System Architecture}
The system is organized as a hierarchy-like graph of homeostatic subsystems, with each of those organized as a hierarchy-like graph of either homeostatic or non-homeostatic subsystems. All subsystems are connected to orchestrate a form of predictive processing\cite{KELLER2018424}.

Each subsystem contains a memory bank that stores a set of events in the form of temporal trace trajectories (see figure~\ref{figure:ttt}). A set of trajectories may represent a homeostatic variable. For instance, the set of trajectories that show core body temperature going from too warm or too cold to nominal temperature represent the core body temperature hvar. 
Temporal traces are used as a means of encoding time. They are computed as a decaying running sum of signals with a decay factor $\delta$. Higher levels in the hierarchy generally use temporal traces with smaller $\delta$ resulting in longer motion trails and trajectories of longer time duration $\tau$ compared to lower levels.

Discussion of specific organization of individual subsystems and modules is beyond of scope of this article.
% We are agnostic about how the temporal trace trajectory memory is implemented.

% The system works with trajectories, not hvars, because an hvar is a human-understandable interpretation of a set of trajectories. It is okay for some trajectories in the system to not be interpretable or be part of multiple interpretations. 

\subsection{System Operation}
The system is bootstrapped by adding a set of trajectories to represent known hvars at various levels in the system hierarchy. Over time, other subsystems learn by remembering trajectories that result from the interaction between sensory signals and the bootstrapped trajectories. Once such trajectories are memorized, they may represent new hvars in corresponding subsystems. Thus, the entire system gets wired with higher level hvars that are grounded to the bootstrap hvars.

The system operation depends on predictive processing\cite{KELLER2018424}, evidence accumulation\cite{Shadlen2016} and top-down signal generation\cite{Adams2013}. Further investigation is needed to identify how a system based on a hierarchy of autoassociative memory modules may produce these characteristics. Here we present a sketch of a plausible, although incomplete, mechanism. In a given homeostatic subsystem, at a given moment in time, select a trajectory matching current conditions. Replay this winning trajectory through top-down signalling over next $T$ time steps. If multiple trajectories are matched with no clear winner, then replay top few matching trajectories over the next few cycles of $T$ time steps each. Repeat till a clear winner emerges by adjusting trajectory strength temporarily through a short term plasticity mechanism.

Confident recall of a higher level trajectory represents an instantiated goal. When such a trajectory is replayed to downstream subsystems, it encourages the whole hierarchy to act according to that goal. Such long term goal trajectories represent cognitive hvars and thus, planned behavior can be interpreted as reflexive behavior corresponding to these higher level cognitive hvars, blurring the boundaries between adaptive, reflexive behavior and intentional, planned behavior.

Cycling through potential future trajectories is seen in the theta band activation of rat hippocampus\cite{Kay2020}. This approach also aligns with evidence accumulation\cite{Shadlen2016} and action planning\cite{Martiny-Huenger2017} through simulated playback from memory. What modules in the system participate in this mechanism remains an open question.

% \subsection{Rewards}
Unlike reinforcement learning, we propose that using an explicit reward signal is not necessary. The more homeostatically helpful a particular trajectory, the stronger it is imprinted in the memory. High utility trajectories 
% are stored with high precision and 
exert a stronger influence on downstream subsystem behavior when replayed. Note that cognitive homeostasis allows hierarchical reinforcement learning\cite{barto2003recent} to emerge within certain subsystems. If a subsystem learns a cognitive hvar that represents an RL-like reward, trajectories for such hvars would be action policies that maximize corresponding rewards.

\subsection{Implementation}
Here we do not propose a specific implementation strategy. Instead, we lay out important characteristics that an implementation must have. The system
1) should preferentially remember favorable trajectories,
% similar to Reinforcement Learning over trajectories\cite{doerr2019trajectory}, 
2) is able to autoassociatively\cite{gritsenko2017neural} recall and replay matching trajectories based on the current state of the system, 
% 3) is built hierarchically where bottom-up signals include sensory signals based on downstream hvars and top-down signals are the most favorable potential continuations of the signal stream, i.e. expected or target future sensory signals, which can be interpreted as predictive processing \cite{KELLER2018424}, 
3) is wired to perform predictive processing\cite{KELLER2018424},
4) has an attention-like mechanisms to identify, instantiate and eventually discard goal trajectories and corresponding cognitive homeostatic variables and 
5) has the ability to learn using sequential, non I.I.D., data continuously without catastrophic forgetting, such as using meta-learning approaches\cite{javed2019meta} to continual learning.

\section{Concluding remarks}
In this article, we described Cognitive Homeostasis, where cognition is interpreted as a homeostatic process, operating over Cognitive Homeostatic Variables. We then described Cognitive Homeostatic Agents that are built as homeostatic hierarchies based on the Cognitive Homeostasis principles.

Much work is needed to explore various aspects of this approach. Is temporal trace trajectory the appropriate unit of memory? What computational framework is suitable for implementing autoassociative trajectory memory? Does the playback of future trajectories use a synchronized clock across various modules? How to integrate bottom-up sensory evidence, the winning trajectories being played back and top-down expectations? How are transient cognitive hvars instantiated? How would planning, curiosity, creativity, emotions and other higher cognitive features emerge? Is this approach biologically plausible, and if so, what are the predictions of this approach that can be investigated through cognitive neuroscience and psychology? Can Cognitive Homeostasis provide normative mechanisms for episodic memory\cite{episodic-1983-tulving} and episodic-like memory\cite{episodic-like-1998-clayton1998}? What does this approach imply for free-will, self-hood and consciousness? Also, a mathematical description based on the Free Energy Principle\cite{Karl2012FEP} formulation would provide formal grounding for cognitive homeostasis. Finally, implementing cognitive homeostatic agents would provide an empirical validation of this approach.

We are cautiously optimistic that this approach would lead to a fruitful line of investigation towards building sophisticated artificial agents. Some of the reasons for our optimism are -  1) Temporal trace trajectories as the unit of memory aligns with engram as a neural memory unit\cite{Josselyneaaw4325}. 2) Mechanism for sampling alternate futures\cite{chadwick2016flexible} as a means of planning may enable planning in physical and conceptual\cite{Constantinescu2016} spaces. 3) All reward-like signals in the proposed system are internal and grounded on the initial set of homeostatic variables, which obviates the need to define explicit rewards or arbitrary mechanisms to derive rewards. 4) We hypothesize that symbols and symbolic manipulation may emerge through the same mechanisms that instantiate and manage cognitive hvars. 5) our approach produces a sparse factor graph\cite{bengio2017consciousness} over events that may result in causal reasoning capabilities. 6) Each homeostatic subsystem behaves in "System 1" \cite{kahneman2011thinking} mode when prediction errors are small. Otherwise it behaves in "System 2" \cite{kahneman2011thinking} mode, where alternate futures need to be sampled over an extended time period in an attempt to minimize prediction errors. 7) Using homeostasis provides a natural and grounded approach to feelings and emotions.

\section{Acknowledgments}
We would like to thank the careful and constructive anonymous reviewers, as well as Lance Hughes for comments and constructive criticism. This work is supported by i3ai.org.

%%%%%%%%%%%%%%%%%%%%%%%%%%%%%%%%%%%%%%%%%%%%%%%%%%%%%%%%%%%%%%%%%%%%%%%%

%%% The next two lines define, first, the bibliography style to be 
%%% applied, and, second, the bibliography file to be used.
\balance
\bibliographystyle{ACM-Reference-Format} 
\bibliography{paper}

%%%%%%%%%%%%%%%%%%%%%%%%%%%%%%%%%%%%%%%%%%%%%%%%%%%%%%%%%%%%%%%%%%%%%%%%

\end{document}